\documentclass{article}
\usepackage{spconf,amsmath,epsfig}
\usepackage{graphicx}
\usepackage{amssymb}
\usepackage{multirow}
\usepackage{makecell,rotating}
\usepackage[english]{babel}

\begin{document}\sloppy

\def\x{{\mathbf x}}
\def\L{{\cal L}}

\title{Multi-Granularity Reasoning for Social Relation Recognition from Images}
%
\name{Meng Zhang$^{1}$, Xinchen Liu$^{2}$, Wu Liu$^{2}$, Anfu Zhou$^{1}$, Huadong Ma$^{1}$, Tao Mei$^{2}$}
\address{$^{1}$Beijing Key Laboratory of Intelligent Telecommunication Software and Multimedia\\
Beijing University of Posts and Telecommunications, Beijing 100876, China\\
$^{2}$JD AI Research, JD.com, Beijing 100101, China
}
%
%

\maketitle

\begin{abstract}
Discovering social relations in images can make machines better interpret the behavior of human beings.
However, automatically recognizing social relations in images is a challenging task due to the significant gap  between the domains of visual content and social relation.
Existing studies separately process various features such as faces expressions, body appearance, and contextual objects, thus they cannot comprehensively capture the multi-granularity semantics, such as scenes, regional cues of persons, and interactions among persons and objects.
To bridge the domain gap, we propose a Multi-Granularity Reasoning framework for social relation recognition from images.
The global knowledge and mid-level details are learned from the whole scene and the regions of persons and objects, respectively.
Most importantly, we explore the fine-granularity pose keypoints of persons to discover the interactions among persons and objects.
Specifically, the pose-guided Person-Object Graph and Person-Pose Graph are proposed to model the actions from persons to object and the interactions between paired persons, respectively.
Based on the graphs, social relation reasoning is performed by graph convolutional networks.
Finally, the global features and reasoned knowledge are integrated as a comprehensive representation for social relation recognition.
Extensive experiments on two public datasets show the effectiveness of the proposed framework.
\end{abstract}
%
\section{Introduction}
Social relation is the close association between individual persons and forms the basic structure of our society.
Recognizing social relations between persons in images can empower intelligent agents to better understand the behaviors or emotions of human beings.
The task of image-based social relation recognition is to classify a pair of persons in an image into one of pre-defined relation types such as friend, family, etc~\cite{SunSF17}.
It has many important applications such as personal image collection mining~\cite{WangGLF10} and social event understanding~\cite{RamanathanY013}.

\begin{figure}[t]
  \centering
  {\includegraphics[width=0.95\columnwidth]{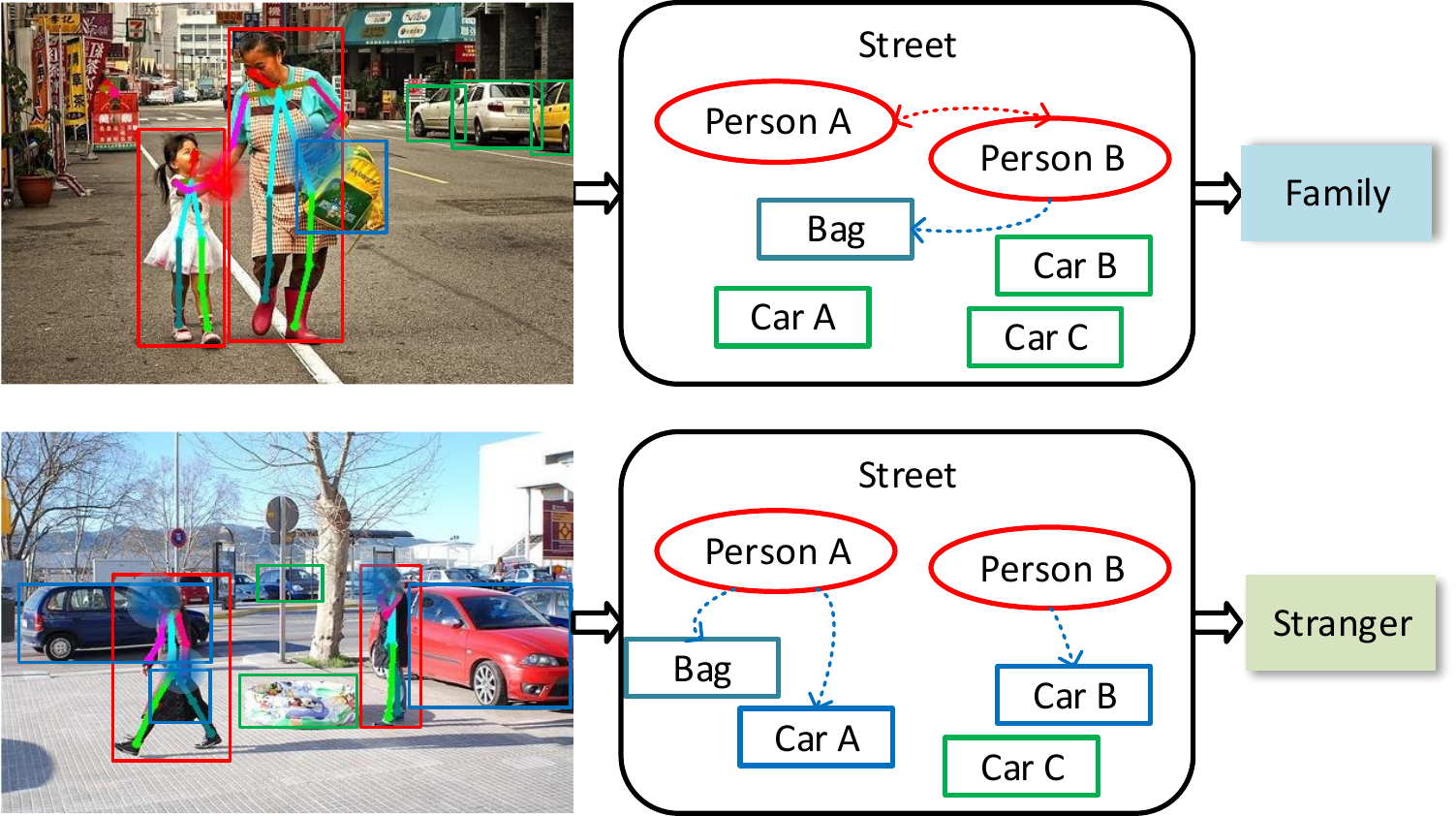}}
  \caption{How do we recognize two persons are family or strangers from an image? The scenes, appearance of persons, and interactions among persons and contextual objects are significant cues for recognition.}
  \label{fig:figure1}
\end{figure}

Sociological analysis based on images has been a popular topic during the last decade~\cite{WangGLF10, YuLPK09}.
Differently, explicit recognition of social relation from images just attracts researchers in recent years with the release of large-scale datasets~\cite{gan2016learning, SunSF17, LiWZK17}.
From traditional hand-crafted features to recent deep learning-based algorithms, the performance of methods achieves significant improvement~\cite{WuDMT09, gan2015devnet, WangCRYCL18}.
However, social relation recognition from images is still a non-trivial problem which faces several challenges.
First of all, there is a large domain gap between visual features and social relations.
It is difficult to recognize the social relation between a pair of person only based on individual visual cues such as scenes, the appearance of persons, or the existence of contextual objects.
Moreover, the scenes and backgrounds of images are varied and wild, which makes global cues uncertain and noisy.
Furthermore, the varieties from the appearance of persons and objects make it difficult to distinguish social relations even for human beings.

Existing methods for social relation recognition usually utilize low-level visual features such as the appearance of persons, face attributes, and contextual objects~\cite{SunSF17, ZhangLLT15}.
Although some approaches explore the relations between persons and objects, they only consider the co-existence in an image~\cite{LiWZK17, WangCRYCL18, gan2016webly}.
However, only depending on the single-granularity representation can hardly overcome the domain gap between visual features and social relations.
How do human beings recognize the relation between two persons in one image?
As shown in Figure~\ref{fig:figure1}, we should consider multi-granularity information including the scene-level knowledge, the appearance of persons, the existence of contextual objects, and the interactions among persons and objects.
Therefore, the multi-granularity reasoning can effectively bridge the domain gap between visual features and semantic social relations.

To this end, we propose a Multi-Granularity Reasoning (MGR) framework for social relation recognition from images.
For global granularity, we adopt a deep Convolutional Neural Network (CNN) to take the whole image as the input for the representation of scenes.
For middle granularity, we adopt an object detector to locate the regions of persons and objects in images, and extract their visual features learned by a CNN as the mid-level knowledge.
For fine granularity, we exploit the pose keypoints of the human body to construct the correlation among persons and objects.
In particular, we not only design a Person-Object Graph (POG) to model the actions from persons to objects, but also build a Person-Pose Graph (PPG) to capture the interaction between paired persons.
Based on these graphs, the Graph Convolution Networks (GCNs) are explored to perform reasoning from the visual domain to social relations by a message propagation mechanism.
Finally, our MGR framework achieves social relation recognition from the complementary information of global knowledge, regional features, and the interactions among persons and objects.

In summary, the contributions of this paper include:
\begin{itemize}
  \item We propose a MGR framework to bridge the domain gap between visual features and social relation for social relation recognition in images.
  \item We design a POG to represent actions from persons to objects, and a PPG to model the interactions between persons guided by human pose.
  Reasoning from visual domain to semantic domain is performed on the graphs by GCNs.
  \item Our framework achieves the state-of-the-art results on two public dataset, which demonstrates the effectiveness of our method.
\end{itemize}

\subsection{Related Work}
The interdisciplinary research of multimedia and sociology has been studied for many years~\cite{WangGLF10, ZhangLLT18}.
Popular topics include social networks discovery~\cite{DingY11}, key actors detection~\cite{RamanathanHAG0F16}, group activity recognition~\cite{BagautdinovAFFS17}, and so on.
In recent years, social recognition from images has attracted attention from researchers~\cite{SunSF17, LiWZK17, WangCRYCL18, ZhangLLT15}.
For example, Zhang~\emph{et al.} proposed to learn social relation traits from face images by CNNs~\cite{ZhangLLT15}.
Sun~\emph{et al.} proposed a social relation dataset based on the social domain theory~\cite{bugental2000acquisition} and exploited CNNs to recognize social relations from a set of attributes~\cite{SunSF17}.
Li~\emph{et al.} proposed to an attention-based dual-glance model for social relation recognition, in which the first glance extracted features from persons and the second glance focused on contextual cues~\cite{LiWZK17}.
Wang~\emph{et al.} proposed to model persons and objects in an image as a graph and perform relation reasoning by a Gated Graph Neural Network~\cite{WangCRYCL18}.
However, they only considered the co-existence of persons and objects in a scene but neglected global information and interactions among persons and objects that are important knowledge for social relation recognition.
Therefore, we propose a Multi-Granularity Reasoning framework to explore complementary cues for social relation recognition.

Graph has been widely adopted to model the visual relations in computer vision and multimedia tasks~\cite{LinWYL15, LiuJLC11, gan2016recognizing, gan2018geometry}.
In recent years, researchers have studied message propagation in graphs by trainable networks, such as Graph Convolutional Networks (GCN)~\cite{kipf2017semi}.
Most recently, these models have been adopted to computer vision tasks~\cite{LiangSFLY16, QiLJFU17}.
For example, Liang~\emph{et al.} proposed a Graph Long Short-Term Memory to propagate information in graphs built on super-pixels for semantic object parsing~\cite{LiangSFLY16}.
Qi~\emph{et al.} proposed a 3D Graph Neural Network to build a k-nearest neighbor graph on 3D point cloud and predict the semantic class of each pixel for RGBD data~\cite{QiLJFU17}.
Inspired by above studies, we propose to represent the interactions of persons and objects as graphs guided by pose keypoints of persons.
Social relation reasoning is performed by GCNs on the built Pose-Guided Graphs.

\begin{figure*}[t]
  \centering
  {\includegraphics[width=\textwidth]{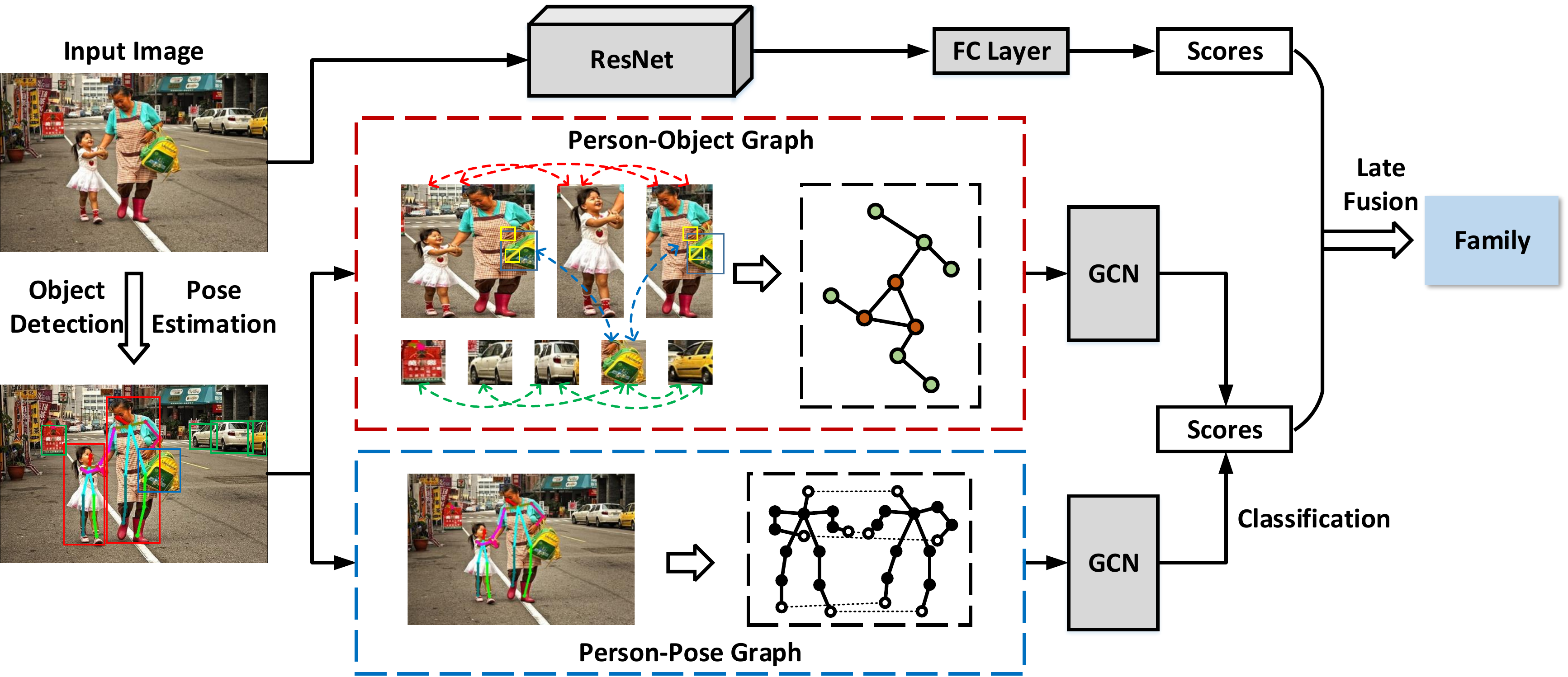}}
  \caption{The overall framework of the Multi-Granularity Reasoning framework.}
  \label{fig:figure2}
\end{figure*}

\section{The Proposed Framework}
\subsection{Overview}
As shown in Figure~\ref{fig:figure2}, the overall structure of the Multi-Granularity Reasoning framework has two main branches.
One branch is designed to learn global features from the whole image.
We adopt a deep CNN, i.e., ResNet~\cite{HeZRS16} to learn knowledge about the scenes for social relation recognition.
The other branch is focused on regional cues and fine interactions among persons and contextual objects for social relation reasoning.
This branch contains three main procedures.
Firstly, we adopt an object detection model, i.e., Mask R-CNN~\cite{HeGDG17}, to crop persons and objects in an image.
The features of each person and object are extracted by CNNs as the regional representation.
Then we adopt the state-of-the-art human pose estimation method~\cite{XiaoWW18} to obtain the keypoints of each person.
After that, we design a Person-Object Graph (POG) to connect each person with not only the other person but also the object that has intersection with the keypoints of the person.
We also build a Person-Pose Graph (PPG) to represent the interaction between two persons based on the keypoints of each person.
Social relation reasoning is performed on the two graphs by graph convolutional networks.
Finally, the social relation between a pair of persons is predicted by integrating the global feature from CNN and the reasoning feature from the GCNs.
Next, we present the details of each branch and procedure.

\subsection{Global Feature Learning}
As mentioned before, the scenes and backgrounds in images can reflect significant information for social relation recognition.
For example, it is more likely that two persons are colleagues if they are in an office or meeting room.
Therefore, we adopt a deep convolutional neural network, i.e., ResNet~\cite{HeZRS16} to learn the visual features from the global view.
In out implementation, we use the 101-layer ResNet, i.e., ResNet-101, as the backbone network due to its powerful capability for representation learning from large-scale data.

\subsection{Pose-Guided Graph Reasoning Branch}
\label{sec:graph}
There are several strategies to model the interactions among persons and objects.
One simple method is focused on the co-existence of persons and objects in an image by combining their features to learn a uniform representation~\cite{LiWZK17}.
Another strategy is to build a graph, in which persons and objects are fully connected~\cite{WangCRYCL18}.
However these approaches are all neglect the interactions among persons and objects.
For example, two persons may have the intimate relation if they walk hand in hand and may be strangers if they stand separately and face to different directions, as shown in Figure~\ref{fig:figure1}.
To connect the correlation between two persons as well as between each person and objects, we propose two graphs respectively.
One is the Person-Object Graph (POG) to model the actions from a person to contextual objects.
The other is the Person-Pose Graph (PPG) to capture the interactions between paired persons.
As mentioned above, we use the bounding boxes of persons and objects, $P = \{p_1, p_2, ..., p_N\}$ and $O = \{o_1, o_2, ..., o_M\}$ from the input image as the nodes.
Moreover, we exploit the pose keypoints of each person to guide the construction of the graphs.

\textbf{Person-Object Graph.}
Persons are the most important factor for social relation recognition.
To learn regional features for persons, the bounding boxes and union areas of two persons are adopted as nodes which are fully connected in the POG.
Besides, the objects can reflect the characters of the scene in an image.
Connecting objects in the scene to each other helps to discover the regional features for social relation recognition.
More importantly, the actions from a person to contextual objects can indicate the role of the person, which are significant cues for social relation recognition.
For example, if two persons are operating a computer, they are more likely to be the professional relation.
Therefore, we use the intersection of a pose keypoint and the object to establish a connection from the person to the object.

The POG is represented as an adjacent matrix, $A_o \in \mathbb{R}^{(N+M) \times (N+M)}$, to model the above interactions among persons and objects.
Since there are two persons and one union of them, the number of person nodes in POG is three.
For person-object connections, let $p_i$ and $o_j$ be the person node and object node, we set
\begin{equation}\label{equ1}
A_o(p_i, o_j) =\left\{
\begin{array}{rcl}
1 & & {IoU(p_i(k), o_j) > 0},\\
0 & & othersise,
\end{array} \right.
\end{equation}
where $IoU(\cdot, \cdot)$ is the Intersection over Union of two bounding boxes, $p_i(k)$ is the $k$-th keypoints of person $p_i$.
The connections between all person nodes and between all object nodes are set to $1$.

\textbf{Person-Pose Graph.}
The human skeleton is a natural graph structure.
For each human skeleton, we establish the association according to the natural connection of the keypoints.
We divide all keypoints into active nodes and passive nodes, where the active nodes include the nose, wrists, ankles.
In our framework, there are $K$ keypoints for each person skeleton.
Therefore, for a pair of two persons in an image, the Person-Pose Graph (PPG) has $2 \times K$ nodes, which can be represented as the adjacent matrix, $A_p \in \mathbb{R}^{(2K) \times (2K)}$.
Let $p_1$ and $p_2$ be the two persons, we first connect the keypoints of the same person following the rule defined in MS-COCO dataset.
The corresponding elements in $A_p$ is set to $1$.
For the inter-person connection, we connect the active nodes of one person to all nodes of the other person and viceversa.
The corresponding elements in $A_p$ is set to $2 - dist(p_1(k), p_2(k'))$, where $dist(\cdot, \cdot)$ is the normalized Euclidean distance between a pair of keypoints on two persons.

To this end, the POG and PPG are built to represent the correlations among persons and objects in images including the interactions between persons, and the actions from persons to objects.
Next we present how to perform social relation reasoning from visual features embedded in the graphs by the Graph Convolutional Network.

\subsection{Reasoning by Graph Convolutional Network}
\label{sec:gcn}
Traditional CNN usually applies 2-D or 3-D filters on images to abstract visual features from low-level space to high-level space~\cite{HeZRS16}.
In contrast, Graph Convolutional Network (GCN) performs relational reasoning by performing message propagation from nodes to its neighbors in the graphs~\cite{kipf2017semi}.
Therefore, we can apply GCNs on POG and PPG to achieve social relation reasoning.

As in~\cite{kipf2017semi}, given a graph with $N$ nodes in which each node has a $d$-length feature vector, the operation of one graph convolution layer can be formulated as:
\begin{equation}\label{equ3}
X^{(l+1)} = \sigma (\tilde{D}^{-\frac{1}{2}}\tilde{A} \tilde{D}^{-\frac{1}{2}} X^{(l)} W^{(l)} ),
\end{equation}
where $\tilde{A} \in \mathbb{R}^{N \times N}$ is the adjacent matrix of the graph, $\tilde{D} \in \mathbb{R}^{N \times N}$ is the degree matrix of $\tilde{A}$, $X^{(l)} \in \mathbb{R}^{N \times d}$ is the output of the $(l-1)$-th layer, $W^{(l)} \in \mathbb{R}^{d \times d'}$ is the learned parameters, and $\sigma (\cdot)$ is a non-linear activation function like ReLU.
In particular, in our social relation reasoning framework, the adjacent matrixes of POG and PPG are $A_o$ and $A_p$ as defined in Section~\ref{sec:graph}.
For a multi-layer GCN, $X^{(0)}=[f(x_1), f(x_2), ..., f(x_N)]^T$ is the initial features for input.
For POG, $f(x_i)$ is the feature vector extracted from the each person or object in the image.
For PPG, the features are obtained from each keypoint of two persons.
Each vector consists of the positions and confidence scores of top 30 points on the heatmap of the keypoint.
The final outputs of GCNs are updated features of nodes, $X^{(L)}$, in the graphs, which are aggregated into an image-level feature vector for social relation prediction.
At last, the scores of GCN and CNN are combined by weighted fusion for the final result.

\section{Experiments}
\subsection{Datasets and Implementation Details}

\textbf{Datasets.}
We conduct experiments on two large-scale social relation datasets, i.e., the People in Social Context (PISC) dataset~\cite{LiWZK17} and the People in Photo Album (PIPA) dataset~\cite{SunSF17}.
The PISC dataset contains several common social relations in daily life, which has a hierarchy of three coarse-level relations and six fine-level ships.
In our experiments, we aim to recognize the six fine-level relations, i.e., friend, family, couple, professional, commercial, and no relation.
We adopt the setting as in~\cite{LiWZK17}, which uses 55,400 instances in 16,828 images for training, 1,505 instances in 500 images for validation, and 3,961 instances in 1,250 images for testing.
The top-1 classification accuracy on each type and the overall mean Average Precision (mAP) are used to evaluate all methods.
The PIPA dataset is annotated based on the social domain theory~\cite{bugental2000acquisition}, in which social life is partitioned into five domains and 16 social relations.
We consider the 16 social relations for recognition in our experiment.
As in~\cite{SunSF17}, the PIPA dataset is divided into 13,729 person pairs for training, 709 for validation, and 5,106 for testing.
The accuracy over all categories is measured in experiments.

\textbf{Construction of Graphs.}
Here we present the nodes of graphs and the features for graph convolution.
The person nodes in POG are the person pair and the union as in~\cite{LiWZK17}.
Five objects boxes detected by Mask R-CNN with high confidence are preserved as objects node.
We use ResNet-101 trained on person images to extract 2048-D features for person nodes and ResNet-101 trained on ImageNet to extract 2048-D features for objects nodes.
To build PPG, we use the state-of-the-art pose estimation model on MS-COCO~\cite{XiaoWW18} to obtain 17 skeleton keypoints as the pose nodes for each person.
The feature of each keypoint is a 30-D vector as mentioned in Section~\ref{sec:gcn}.

\textbf{Training of Networks.}
In our framework, the ResNet with attentive module and the GCNs are trained separately.
We train the ResNet-101 network which is pretrained on ImageNet.
The model is first trained by Adam optimization with $lr = 10^{-3}$ then fine-tuned with $lr = 10^{-4}$.
For the GCN propagation model, we use SGD optimization with momentum of 0.9.
The $lr$ starts from 0.01 and multiplies 0.1 by every 20 and 30 epochs for PISC dataset and PIPA dataset, respectively.
During testing, the fusion weights for the results of global feature and PGG are 0.4 and 0.6, respectively.

\subsection{Comparison with the state-of-the-art methods}

\begin{table*}[t]
\centering
\caption{Comparison to the state-of-the-art methods on PISC and PIPA datasets.}
\label{tab:table1}
\begin{tabular}{l|c|c|c|c|c|c|c|c}\hline
                & \multicolumn{7}{c|}{PISC}& PIPA \\ \hline
                & \rotatebox{0}{Friend}   & \rotatebox{0}{Family}   & \rotatebox{0}{Couple}   & \rotatebox{0}{Profes.}   & \rotatebox{0}{Commer.}   & \rotatebox{0}{No Rela.} & \rotatebox{0}{mAP} & \rotatebox{0}{Accuracy} \\  \hline
Two stream CNN~\cite{SunSF17}     & - & - & - & - & - & - & - & 57.2 \\
Union-CNN~\cite{LuKBL16}          & 29.9 & 58.5 & 70.7 & 55.4 & 43.0 & 19.6 & 43.5 & - \\
Dual-glance~\cite{LiWZK17}        & 35.4 & \textbf{68.1} & \textbf{76.3} & 70.3 & \textbf{57.6} & 60.9 & 63.2 & 59.6 \\
GRM~\cite{WangCRYCL18}            & 59.6 & 64.4 & 58.6 & 76.6 & 39.5 & 67.7 & 68.7 & 62.3 \\
MGR (ours)                        & \textbf{64.6} & 67.8 & 60.5 & \textbf{76.8} & 34.7 & \textbf{70.4} & \textbf{70.0} & \textbf{64.4} \\ \hline
\end{tabular}
\end{table*}

\begin{table*}[t]
\centering
\caption{Ablation study of our framework on the PISC dataset.}
\label{tab:table2}
\begin{tabular}{l|c|c|c|c|c|c|c}\hline
                     & \rotatebox{0}{Friend}   & \rotatebox{0}{Family}   & \rotatebox{0}{Couple}   & \rotatebox{0}{Profes.}   & \rotatebox{0}{Commer.}   & \rotatebox{0}{No Rela.} & \rotatebox{0}{mAP} \\  \hline
Global        & 55.7            & 55.4          & 48.0          & \textbf{77.0} & \textbf{35.9} & 51.4 & 57.9 \\
POG (w/o pose)& 55.4            & 64.8          & 50.4          & 71.8          & 31.2          & 67.3 & 65.5 \\
POG           & 59.4            & 61.9          & \textbf{62.5} & 68.9          & 31.1          & 63.7 & 66.3 \\ \hline
PGG + PPG     & 62.2            & 64.5          & 61.3          & 70.6          & 31.4          & 68.3 & 68.0 \\
Global + POG  & 63.8            & 65.3          & 60.2          & 76.5          & 34.5          & 67.0 & 69.1 \\ \hline
MGR           & \textbf{64.6}   & \textbf{67.8} & 60.5          & 76.8          & 34.7          & \textbf{70.4} & \textbf{70.0} \\ \hline

\end{tabular}
\end{table*}

To validate the effectiveness of the proposed Multi-Granularity Reasoning framework, we compare it with several state-of-the-art methods on PISC and PIPA.
The details of methods are as follows:

\textbf{1) Two stream CNN~\cite{SunSF17}}.
This method is the baseline on the PIPA dataset.
It is focused on human bodies and head regions.
A group of attributes such as gender, age, clothing, and actions are extracted for social relation recognition.

\textbf{2) Union-CNN~\cite{LuKBL16}}.
This method adopt a CNN model to identify the relationship through persons¡¯ joint area.
The results of Union-CNN are from the implementation in~\cite{SunSF17}.

\textbf{3) Dual-glance~\cite{LiWZK17}}.
This approach designs a deep CNN with attention on regions of persons and objects to improve social relation prediction.

\textbf{4) Graph Reasoning Model (GRM)~\cite{WangCRYCL18}}.
This is the state-of-the-art model for social relation recognition on the datasets of PIPA~\cite{SunSF17} and PISC~\cite{LiWZK17}.
It represents the persons and objects existing in an image as a weighted graph, on which the social relation is predicted by a Gated Graph Neural Network~\cite{LiTBZ15}.

\textbf{5) Multi-Granularity Reasoning (MGR)} is our proposed framework for social relation recognition.

The experimental results are listed in Table~\ref{tab:table1}.
From the results, we can find that Union-CNN and two stream CNN obtain relatively poor accuracy.
This is because they only consider person-related features such as the appearance of bodies, face attributes, clothing, and so on.
These features are insufficient for comprehensively represent the social relation in images.
Moreover, since Dual-glance and GRM explore both persons and contextual objects, they achieve better accuracy than the former two methods.
This shows the role of contextual objects for human social relation recognition.
The Dual-glance method has superior results on family, couple, and commercial, while has worse results on other relations.
This reflects the attention mechanism on contextual objects can effetely discover the key objects but also may be misleaded by unrelated objects for social relation recognition.
Furthermore, our proposed MGR framework not only obtain the best overall accuracy on both PISA and PIPA but also achieves uniform performance on all types of relations.
This demonstrates that the scene-level global knowledge, appearance of persons and objects, and the interactions between persons and objects can provide complementary features for bridging the visual relation and social relation in images.
Therefore, by the integration of the above features, our MGR framework can achieve accurate social relation reasoning on comprehensive information from images.

\subsection{Analysis}
This section explores the role of each module in MGR framework on the PISC dataset.
Table 2 lists following combinations of multi-granularity modules:

\textbf{1) Global} is the global features learned by ResNet.

\textbf{2) POG w/o pose} uses a Person-Object Graph that is built by fully connecting all persons and objects in the graph without considering the pose of each person.

\textbf{3) POG} is social relation reasoning on our Person-Object Graph guided by pose estimation of persons.

\textbf{4) POG + PPG} is the combination of Person-Object Graph and Person-Pose Graph by joint training two GCNs on two graphs.

\textbf{5) Global + POG} adopts the late fusion strategy to predict social relation from global and person-object knowledge.

\textbf{6) MGR} is the complete framework.

For individual modules, POG and PPG outperform Global, which shows that regional features and interactions are more significant than global knowledge.
Nonetheless the global part is better than POG and PPG for professional and commercial.
This indicates that the non-intimate relations are more correlated with scenes.
Moreover, the POG outperforms the POG without pose guidance, which demonstrates the importance of actions from persons to objects.
Furthermore, by integration of different modules, the results are better than individual ones.
The complete MGR achieves the best performance.
The multi-granularity features have complementary effects for social relation recognition especially for the categories of friend, family, and no relation.

However, by the observation on the results, we find that social relation recognition in images is still a very challenging task.
The intimate relations like friends, family, and couple are usually ambiguous under different background and easily confused even for human beings.
Besides, the distribution of different relations in the datasets and even in our daily life is very unbalanced.
Therefore, learning-based methods such as deep CNNs may be failed for the relations with few samples.
In future work, we will explore how to further discover the context cues in the whole scene and overcome the unbalance of samples for social relation recognition.

\section{Conclusion}
In this paper, we propose a MGR framework for social relation recognition from images.
To learn global knowledge, we adopt a deep CNN to take the entire image as the input for training.
For mid features, we extract visual features from bounding boxes of persons and objects.
More importantly, we explore the fine-granularity keypoints of persons to model the interactions amount persons and objects.
In particular, we design a POG to model the actions from persons to contextual objects.
We build the PPG by the keypoint-based inter-person connections to represent the interactions between two persons in an image.
Based on graphs, social relation reasoning is performed with the GCNs by a message propagation mechanism.
At last, our MGR framework achieves social relation recognition from the complementary multi-granularity information.
Extensive experiments on two public datasets show that our framework outperforms the state-of-the-art methods.

{\small
\bibliographystyle{ieeebib}
\bibliography{icme2019abbr}
}

\end{document}